\pdfoutput=1

\documentclass[11pt]{article} 

\usepackage{emnlp2021}

\usepackage{times}
\usepackage{latexsym}
\usepackage{graphicx}
\graphicspath{ {./fig/} }
\DeclareGraphicsExtensions{.pdf,.png}

\usepackage[T1]{fontenc}

\usepackage[utf8]{inputenc}

\usepackage{microtype}

\usepackage{enumitem}
\setitemize{noitemsep,topsep=0pt,parsep=0pt,partopsep=0pt}
\definecolor{cite}{rgb}{0.6,0.6,1.0}
\usepackage{booktabs}
\usepackage{graphicx}
\usepackage{float}
\usepackage{tabularx}
\usepackage{soul}
\usepackage{subcaption}

\newcommand{\toolname}{CARE}

\title{\emph{\hl{\toolname{}}}: Collaborative AI-Assisted Reading Environment}%

\author{Dennis Zyska$^*$, Nils Dycke$^*$, Jan Buchmann, Ilia Kuznetsov, Iryna Gurevych\\ 
Ubiquitous Knowledge Processing Lab (UKP Lab)\\
Department of Computer Science and Hessian Center for AI (hessian.AI)\\
Technical University of Darmstadt \\
\texttt{ukp.informatik.tu-darmstadt.de}}

\begin{document}
\maketitle

\def\thefootnote{*}\footnotetext{These authors contributed equally to this work}\def\thefootnote{\arabic{footnote}}

\begin{abstract}

Recent years have seen impressive progress in AI-assisted writing, yet the developments in AI-assisted \emph{reading} are lacking. We propose \emph{inline commentary} as a natural vehicle for AI-based reading assistance, and present CARE: the first open integrated platform for the study of inline commentary and reading. CARE facilitates data collection for inline commentaries in a commonplace collaborative reading environment, and provides a framework for enhancing reading with NLP-based assistance, such as text classification, generation or question answering. The extensible behavioral logging allows unique insights into the reading and commenting behavior, and flexible configuration makes the platform easy to deploy in new scenarios. To evaluate CARE in action, we apply the platform in a user study dedicated to scholarly peer review. CARE facilitates the data collection and study of inline commentary in NLP, extrinsic evaluation of NLP assistance, and application prototyping. We invite the community to explore and build upon the open source implementation of CARE\footnote{\url{https://github.com/UKPLab/CARE}}.

\end{abstract}

\section{Introduction}
\label{sec:intro}

Individual and collaborative text work is at the core of many human activities, including education, business, and research. Yet, reading text is difficult and takes considerable effort, especially for long and domain specific texts that require expert knowledge. While past years have seen great progress in analyzing and generating text with the help of AI -- culminating in strong generative models like GPT-3 \cite{gpt3} and ChatGPT \cite{instructGPT}\footnote{\url{https://openai.com/blog/chatgpt}} -- the progress in applications of AI to reading and collaborative text work is yet to match these achievements. The ability of modern generative models to create natural-sounding but factually flawed outputs \cite{ji2022survey} stresses the need for supporting humans in critical text assessment. %

\begin{figure}[t]
  \centering
  \includegraphics[width=8.3cm]{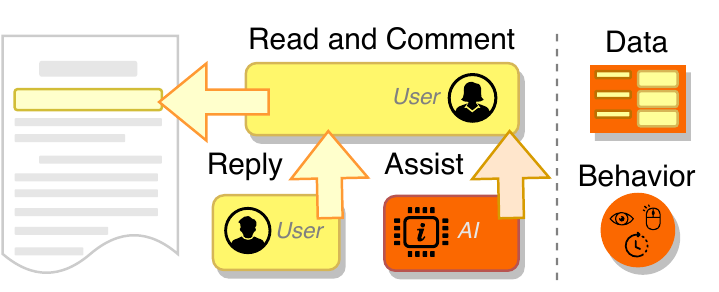}
  \caption{CARE allows users to collaboratively read and discuss texts, provides a generic interface for AI-based reading assistance, and collects research-ready textual and behavioral data.}
  \label{fig:architecture}
\end{figure}

Humans use annotations to read and collaborate over text, from hand-written print-out notes to highlights in collaborative writing platforms. This makes in-text annotations -- \textit{inline commentaries} -- a promising vehicle for AI-based reading assistance. For example, an AI assistant could automatically classify the user's commentaries, or verify and provide additional information on the highlighted passages. Yet, the lack of data and key insights limits the NLP progress in this area: from the foundational perspective, we lack knowledge about the language of inline commentaries, as most of this data is not openly available for research. From the applied perspective, little is known about the hands-on interactions between humans and texts, how they translate into NLP tasks, and how the impact of NLP-based assistance on text comprehension can be measured. While ethical, controlled data collection has been receiving increasing attention in the past years \cite{texprax}, data collection tools for inline commentary are missing, and so are the tools for applying and evaluating NLP models within a natural reading environment.

To address these limitations, we introduce CARE: a \textbf{C}ollaborative AI-\textbf{A}ssisted \textbf{R}eading \textbf{E}nvironment, where users can jointly produce inline commentaries on PDF documents in an intuitive manner, connected to a model-agnostic and flexible AI assistance interface.
Unlike existing labeling tools, CARE provides a (1) familiar, task-neutral environment for collaborative reading similar to the tools used in everyday text work; unlike off-the-shelf reading and writing applications, CARE offers (2) structured machine-readable data export functionality, including both inline commentary and behavioral data; unlike task-specific AI-assisted reading tools, CARE features a (3) generic interface for integrating NLP modules to support reading and real-time text collaboration.

Our contribution has multiple audiences. For NLP researchers, CARE makes it possible to efficiently collect inline commentary data in a standardized manner, and provides a generic interface for the extrinsic evaluation of NLP models. For application designers, CARE offers an extensible platform and behavioral metrics to study how humans interact with texts. For users, CARE enables the development of new, innovative applications built around AI-assisted reading such as interactive e-learning, community-based fact-checking, and research paper assessment. To foster the progress in AI-assisted reading research, we make the implementation openly available and easy to deploy and to extend.

\section{Background}
\label{sec:relw}

\begin{figure}
  \centering
  \includegraphics[width=0.7\linewidth]{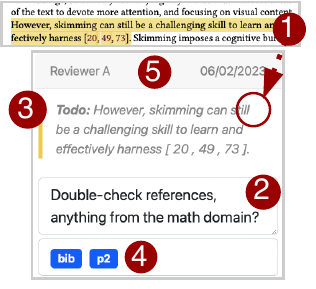}
  \caption{An inline commentary in CARE consists of a highlight (1) optionally associated with a commentary text (2), a label (3), a number of free-form tags (4) and metadata (5), e.g. user name and creation time.}
  \label{fig:inlinecomm}
\end{figure}

\subsection{Terminology and Requirements} 
The term "annotation" allows for broad interpretation and encompasses both the results of controlled annotation studies that enrich text with a specific new information layer (e.g. named entities), and the less-regulated, natural annotations that humans produce when they work with text. Yet, the two annotation mechanisms are fundamentally different. Labeled NLP data is usually obtained via annotation studies -- supervised campaigns that involve formalized tasks and labeling schemata, detailed guidelines, and are supported by specially designed annotation software that requires training of the annotators. However, collecting natural annotation data requires the opposite: the process should minimally interfere with the user's workflow, and the tool should provide a natural environment for working with text given the task at hand. Our work addresses annotation in the latter sense. To avoid ambiguity, we propose the term \emph{inline commentary} to denote in-document highlights left by the users while reading and collaborating on text, potentially associated with commentary text, tags and metadata (Figure \ref{fig:inlinecomm}). We reserve the term \emph{labeling} for the traditional NLP markup. 

With this distinction in mind, for a tool to support the study of inline commentary we define the following \textbf{requirements} distributed among the key \textsc{user groups}:

\textbf{A. Natural environment:} The tool should provide the \textsc{reader} with a natural reading environment, specified as allowing the \textsc{reader} to (A1) leave inline commentaries on the documents in (A2) common reading formats like PDF, while requiring (A3) minimal to no training.

\noindent
\textbf{B. Collaboration:} The tool should run (B1) online and support (B2) real-time collaboration where the \textsc{readers} can leave, see and reply to each others' commentaries in an on-line fashion.

\noindent
\textbf{C. Data management:} The tool should enable \textsc{researchers}, \textsc{application developers} and \textsc{administrators} to easily (C1) import new documents, (C2) collect inline commentary and \textsc{user} behavior data, and (C3) export this data in a machine-readable format for further scrutiny.

\noindent
\textbf{D. Openness and extensibility:} Both documents and inline commentaries might contain confidential data. It is thus crucial that a tool can be (D1) self-hosted on-premise and allows controlling user access to the data. AI-assisted reading has many potential use cases, stressing the need for (D2) high configurability and easy deployment of the tool. To promote transparency and facilitate extension, the platform should be available as (D3) open-source.

\noindent
\textbf{E. AI assistance:} Finally, the tool should provide an easy way to (E) integrate AI assistance modules for \textsc{researchers} and \textsc{developers} to support \textsc{users} in reading and comprehending text.

\subsection{Related tools}

We identify four broad groups of software tools falling within the scope of our requirements, which we briefly exemplify below.
Our overview demonstrates the wide use of inline commentary "in the wild" and underlines the limitations of the existing solutions for the systematic study of inline commentary and AI-assisted reading in NLP.

\paragraph{Readers} Highlighting and inline commentary are core features of most standalone reading tools, from PDF viewers
like Adobe Acrobat Reader\footnote{\url{https://www.adobe.com/acrobat/pdf-reader.html}} to literature management software like Mendeley\footnote{\url{https://www.mendeley.com}}.
The most commonly used tools are proprietary and thereby hard to extend, and do not offer management, collection and export of fine-grained data, making them unsuitable for the study of inline commentary. 
While a few dedicated reading applications like ScholarPhi \cite{scholarphi}, Scim \cite{fok2022scim}, SciSpace\footnote{\url{https://typeset.io}}, and Scrible\footnote{\url{https://www.scrible.com}} do offer machine-aided reading assistance, they focus on their particular use cases, lack data collection functionality and extensibility, and can not be easily hosted on-premise to protect potentially sensitive user and document data. 

\paragraph{Social annotation} Focusing on the collaborative aspect of reading, social annotation platforms %
allow users to exchange their inline commentaries via a centralized platform. A prime example is Hypothes.is\footnote{\url{https://web.hypothes.is}}, which offers a natural environment, is available open-source and provides a standardized mechanism for exporting inline commentary. Yet, the platform is not easy to extend and customize, and does not offer a standardized mechanism for integrating AI-assistance or behavioral data collection. While not being based on Hypothes.is, CARE adopts many of its design ideas, including the appearance and functionality of the annotation sidebar, %
utilities to locate inline commentaries in the document text, as well as the underlying data structure of the annotations.

\paragraph{Authoring tools} Inline commentary is featured in many text authoring tools, from standalone office applications like Microsoft Office\footnote{\url{https://www.office.com/}}
to collaborative web-based platforms like Google Docs\footnote{\url{https://www.google.com/docs/about}} 
and Overleaf\footnote{\url{https://www.overleaf.com}}. While widely used and familiar, these applications are hard to tailor to the needs of a particular scientific study, offer limited data export capabilities, lack flexible AI integration for assistance, and are either implemented as standalone desktop applications (impeding real-time collaboration), or do not allow self-hosting, making ethical data collection and storage challenging.

\paragraph{Labeling tools} The rapid progress in NLP of the past decades has been accompanied by the evolution of general-purpose tools used to acquire labeled data \cite{Neves2019}, from early desktop applications like \textit{WordFreak} \cite{morton2003} to modern extensible, web-based, open-source environments like \textit{brat} \cite{brat2012}, \textit{labelstudio}\footnote{\url{https://labelstud.io}}, \textit{docanno}\footnote{\url{https://github.com/doccano/doccano}} and \textit{INCEpTION} \cite{klie2018}. CARE inherits many concepts from NLP annotation platforms -- including coupling of external recommenders \cite{klie2018}, tag sets and study management functionality, and flexible data export. Although not specifically designed for controlled labeling scenarios, CARE can be used as a lightweight labeling tool with collaboration and assistance capabilities.

\section{Platform Description}
\label{sec:demo}

CARE addresses the gap in existing solutions that prevents the study of inline commentary and AI-assisted reading. %
Here we review the main components of CARE from the user perspective, while the next Section outlines the key technical aspects of our open implementation. We discuss the components in order of importance and refer to the Appendix \ref{app:appdetails} for the illustration of a typical user journey. %

At the core of CARE is the \textbf{reading component} which allows users to attach inline commentaries to documents. To ensure that the visual representation of the document remains true to its source and stable across platforms, CARE focuses on PDF as the main source format\footnote{While support for other document formats is planned, we note that any textual document can be converted into a PDF.}. An inline commentary can amount to a simple highlight attached to a continuous text span, can be associated with a free-text note, and can carry a label from a pre-configured label set, as well as any number of free-form tags (Figure \ref{fig:inlinecomm}). It is possible to add document-level commentaries that are not attached to a span. Inline commentaries are displayed in the dedicated \textbf{CARE sidebar} and can be navigated and edited. The process is \textbf{collaborative}: multiple users can leave inline commentaries on the same document and reply to them in real time. The commentaries are saved and can be revisited at a later point; the resulting data can be \textbf{exported} in an easy-to-use data format, individually or in aggregate, and displayed within the user interface of CARE. In addition to the textual data, CARE collects and exports basic behavioral metrics; for instance, the time of highlight creation and the users' scrolling behavior within the document.

The second key component of CARE is \textbf{AI assistance}: the inline commentary data can be routed to an arbitrary external NLP module, which returns the prediction that can be displayed in the annotation component \emph{in close-to-real-time} as labels, inline commentary replies, or via a custom UI. The interaction between users and AI assistance is mediated by a flexible \textbf{broker} system that distributes the processing tasks among a set of NLP models. Multiple AI assistance model instances can be acting simultaneously, and the pool of models can be extended easily through registering a new model node at the broker backend. At the moment of writing, CARE provides examples to supports integration of any pre-trained model compatible with the \textit{huggingface transformers} API \cite{hf_transformers} by simply changing the configuration parameters. The model then has access to the inline commentary text, highlighted span from the main document, labels, tags and metadata. It is possible to adapt CARE to use models based on other frameworks. 

Finally, CARE features a flexible and configurable \textbf{dashboard} that provides quick access to user and system settings, document and label set management, and study management. In particular, the \textbf{user management} component is responsible for registration, authentication and authorization; to encourage responsible data collection and ensure that the collected inline commentary data can be used in research, CARE features sample \textbf{informed consent} forms that users are presented upon registration, along with the necessary \textbf{licensing disclaimers}, which can be refined by the study administrator.

\section{System Design}
\label{sec:system}

\begin{figure}
  \centering
  \includegraphics[width=0.85\linewidth]{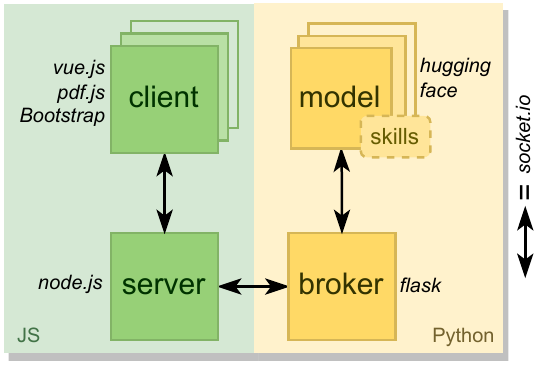}
  \caption{Overview of CARE system architecture.}
  \label{fig:syscomp}
\end{figure}

CARE is designed to be generic, modular and extensible (Figure \ref{fig:syscomp}). The ability to build and deploy CARE via a Docker container makes it easy to set it up in new environments. While CARE features detailed documentation, here we provide a high-level overview of the system design. CARE follows a client-server architecture, preferring client-side operation whenever possible to speed up execution and reduce network traffic and server load. This results in a clear separation of responsibilities between the client, the server and the NLP assistance components of CARE and affords high modularity. While the main client-server pair is purely JavaScript-based, the AI models and the broker are implemented in Python to facilitate the NLP assistance development by the natural language processing community.

The web-based \textbf{CARE client} is responsible for frontend rendering and annotation functionality. The client is fully implemented in \textit{vue.js}\footnote{\url{https://vuejs.org}}, allowing dynamic rendering, modular frontend structure and reuse of original and third-party components.
\textit{Bootstrap}\footnote{\url{https://getbootstrap.com}} is used throughout the frontend to ensure consistent styling and responsive design; document rendering is handled via \textit{pdf.js}\footnote{\url{https://mozilla.github.io/pdf.js}}. In addition, we adopt the localization code from hypothes.is\footnote{\url{https://github.com/hypothesis/client}} to locate inline commentaries in the document.
The \textbf{CARE server}, in turn, is responsible for synchronizing the data among clients, authentication and authorization, and for connecting to external services, including the AI-assistance broker. 
In line with the JavaScript-based frontend, the backend is implemented in \textit{node.js}\footnote{\url{https://nodejs.org/en}} as a cross-platform runtime environment. 
As keeping message transition time low is crucial for collaboration and AI assistance, we base all communication on the \textit{WebSocket protocol}\footnote{\url{https://datatracker.ietf.org/doc/html/rfc6455}}. Persistent bidirectional connection between the client and server components enables real-time exchange of messages and reduces communication time to the possible minimum by reducing the number of connection setups (i.e., three-way handshakes).

AI assistance in CARE is implemented by routing user requests to separately hosted NLP models abstracted into \textbf{Skills}: high-level machine-readable specifications of assistance functionalities including inputs, outputs and model configurations \cite{baumgartner-etal-2022-ukp}. The current implementation of NLP assistance in CARE is built on top of the \textit{huggingface} pipeline, making it easy to integrate a wide range of pre-trained models; we provide sample code to facilitate building self-registering docker containers for NLP model deployment. The interactions between the server and the NLP models is mediated by a \textbf{Broker} system which distributes user requests among NLP models depending on the necessary skill.

\section{User Study}
\label{sec:studies}

To evaluate and refine the reading environment of CARE in the context of a collaborative applied task (requirements A and B), and to ensure the data export functionality (C) and the extensibility (D) of the system, we have extended the base configuration of CARE to accommodate a custom reading scenario and conducted a user study. We describe the core components of the study here and refer to the Appendix \ref{ass:pool} for details.

\paragraph{Task} Scholarly peer review is a prototypical example of close reading accompanied by note-taking, where an expert assesses a manuscript in terms of its originality, readability, validity and impact \cite{jefferson_measuring_2002}. We adopted critical reading that takes place during peer review as a basis for our task. The participants of the study were provided with a manuscript-to-review and instructed to leave self-contained annotations on the manuscript while reading. To incentivize reviewers to perform the task rigorously, we simulated a subsequent acceptance-decision-making phase based on the provided annotations. To support the scenario, we extended CARE to allow reviewer-paper assignment and decision-making functionality.

\paragraph{Study design} We selected two nine-page papers (P1 and P2) from the F1000RD corpus \cite{kuznetsov2022revise}, both dedicated to broad academic topics that are understandable for participants with academic background. Before the study, the participants were instructed about the task, and given 15 minutes to familiarize themselves with the CARE environment. The participants were then split into two groups and assigned paper P1 or P2 based on their group. The participants proceeded to review their assigned paper individually under time constraints (40 minutes), following to the task definition provided above. After the time elapsed, the papers were exchanged between the groups, and the participants were asked to make an acceptance decision for the unseen paper given the inline commentaries produced by a reviewer from the other group. The task was performed in English.

\paragraph{Participants}
In total $11$ researchers from the digital humanities ($6$) and social sciences ($5$) participated in the study. A pre-study questionnaire verified that the participant demographics were diverse and that more than $60\%$ of the researchers were at a post-doctoral or professorial level in their careers with adequate English proficiency.

\begin{figure}[t]
    \centering
    \includegraphics[width=0.85\linewidth]{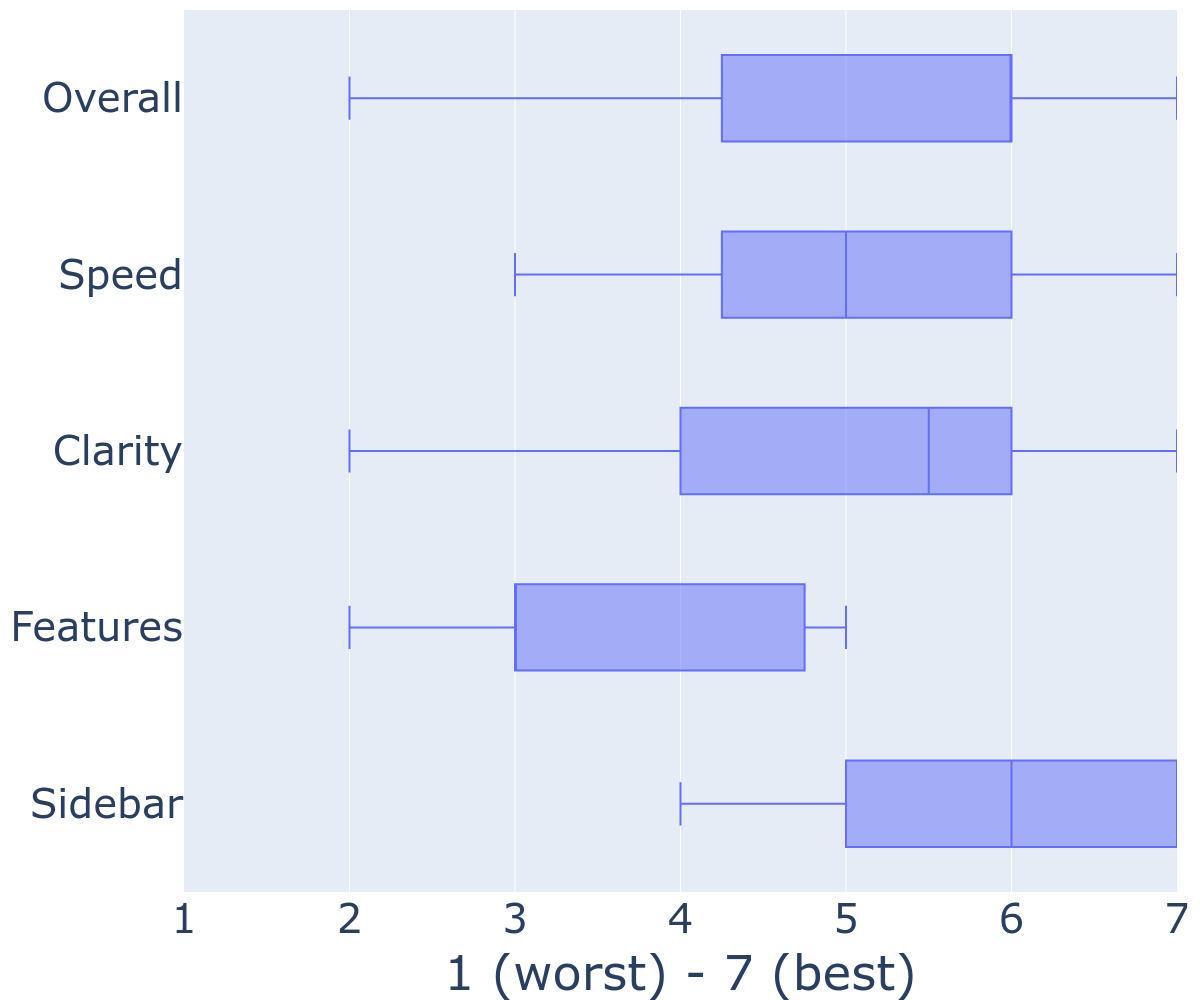}
    \caption{Usability questionnaire results.}
    \label{afig:full_questionaire}
\end{figure}

\paragraph{Usability}
After the study, we conducted a usability survey including a subset of the standardized PSSUQ questionnaire \cite{borsci2015assessing}, as well as free-form questions (details in Appendix  \ref{ass:usability}). As Figure \ref{afig:full_questionaire} shows, the majority of participants were satisfied with using CARE for their task and found that the tool provided adequate speed. Most reported that CARE was clear and easy to use, and appreciated the sidebar functionality. The survey revealed a few feature requests including the ability to arrange inline commentaries in the sidebar by different criteria, and the ability to leave annotations on figure elements.

\paragraph{Data: Inline commentaries}
The export functionality allowed us to examine the data resulting from the study. In total, participants created $200$ inline commentaries of which $151$ were associated with commentary text, $17\pm 7.08$  commentaries per user per document on average. The highlight spans comprise of on average $161 \pm 151.09$ characters and vary vastly from single words up to full paragraphs, selections of two to three sentences being the most common. The associated commentaries have $80 \pm 109.98$ characters on average, ranging from very short remarks of a single word (e.g. "references?", "why?") to full summarizing paragraphs. These results demonstrate the variability of natural inline commentary use.

\paragraph{Data: Reading behavior}
Behavioral metrics integrated into CARE allowed us to observe how the participants used the tool to perform the task at hand. We observed that $35$ annotations ($17.5\%$) were deleted after creation, prompting us to improve the inline commentary edit functionality in the tool; nearly all participants ($70\%$) made use of the ability to quick-scroll from the in-text highlights to the annotations in the sidebar, while the opposite direction (quick-scroll to the highlight \emph{from} the sidebar) was only used rarely. The page tracking functionality allowed insights into how participants assessed the papers: by measuring the time spent on each respective page, we established that the participants spent the least amount of time reading bibliography, whereas method and conclusion sections received most scrutiny. We elaborate on these results in the Appendix \ref{ass:behavior}.

\section{CARE and AI Assistance}

\paragraph{Data collection} CARE enables the collection of inline commentary data that can be used to study inline commentaries and to create new datasets for NLP assistance model development. The collaboration functionality of CARE allows gathering the data about reader interactions within the tool, and the support for free-form tagging and controlled labeling offers great opportunities for collecting user-generated silver data for model pre-training and fine-tuning.

\paragraph{Assisted reading} Out of the box, CARE supports integration of any pre-existing \emph{huggingface transformer} model into the reading workflow, which opens a wide range of possibilities for applying previously developed models "in the wild". To provide feedback to the reader, a pre-trained model can be used to enrich inline commentaries with labels, i.e. prompting the reader to provide additional detail, assessing the politeness \cite{politeness}, specificity \cite{specificity} or sentiment \cite{sentiment} of a commentary. In addition, the power and flexibility of modern generative models like T5 \cite{t5} allow performing a wide range of text-to-text tasks to assist reading, from question answering to summarization of highlighted passages, with the results rendered as automatically generated replies to the user's inline commentaries. The CARE repository provides sample code for NLP model integration.

\paragraph{Extrinsic evaluation} Finally, the behavioral metrics provided by CARE allow to study both how humans read and comment on documents, and how AI assistance impacts this behavior, for example by recording the order in which parts of the document get accessed, or the time needed to create the commentaries. While the current implementation only supports basic time- and location-based measurements, we envision a wide range of extensions that would help us study the impact of AI assistance on reading and text work.

\section{Conclusion and Future Work}
\label{sec:conclusion}

This paper has presented CARE -- a new open platform for the study of inline commentary and AI-assisted reading. CARE enables efficient inline commentary and behavioral data collection for NLP, and supports a wide range of collaborative reading scenarios, while requiring minimal effort to use. The extensible NLP assistance interface allows using CARE for rapid prototyping and extrinsic evaluation of NLP modules that support reading and text-based collaboration. Planned extensions of CARE include support for non-PDF document processing and automatic text highlighting, improved human-in-the-loop functionality and scalability, as well as further development of the onboard behavioral metrics. We invite the community to use our tool and contribute to its further development\footnote{\url{https://github.com/UKPLab/CARE}}.

\section*{Ethics}
\label{sec:eth}

The experiments performed in this study involved human participants who gave explicit consent to the study participation and to the storage, modification and distribution of the collected data. The arbitrary username selection by the users ensured that the behavioral data did not allow any association with the participants unless they decided to reveal this information. We report the demographic distribution of the participants in the Appendix. The documents used in the study are distributed under an open license. Although we have attempted to reflect the reading-for-peer-review workflow as closely as possible, we note that the study might still not be fully representative of the reading practice during peer review, as the participants were strictly limited in time to perform the task, and the selected papers were not necessarily from the participants' domains of specialist expertise.

Any application of AI to assisting humans in performing real-world tasks bears risk. We stress the need to control for bias, harmful content and factuality of the AI models used to assist reading and text work -- especially in the case of large pre-trained generative models. We deem it equally important to educate the users of AI-assisted reading tools about the limitations and risks associated with the integrated assistance models.

From the data collection perspective, we note that all data collected with CARE is human-generated personal data, in particular the behavioral data. We thus \emph{require} the users of the tool to provide explicit informed consent on the data collection upon registration. In addition, the users must explicitly agree with the optional collection of behavioral statistics before any of this data is transferred to the server (opt-in). We stress that while sufficient for controlled studies, in a real application environment these measures would need to be extended by allowing the users to change their decision at a later point and specify the parts of their data that are included into data collection.

From the privacy perspective, CARE allows registration with an arbitrary username, first and last name, e-mail and password. The choice and management of the usernames and user identities are left to the study administrator -- we note that if the usernames are not assigned at random and associated with additional data, this needs to be incorporated into the informed consent form upon registration. CARE implements standard security measures to protect the data, and complete access to the data (documents, inline commentaries, behavioral data) is restricted to the application and server administrator. The security mechanism of the broker and thus of the AI-assistance is currently set via a token defined during the installation of the platform. It is up to the administrator to ensure that the token is kept private, otherwise the models can be used by unwanted users. We stress that for some application scenarios -- e.g. dealing with sensitive or confidential documents or performing advanced behavioral measurements -- additional security measures should be considered to protect the data.

\section*{Acknowledgements}

This work has been funded by the German Research Foundation (DFG) as part of the PEER project (grant GU 798/28-1) and by the German Federal Ministry of Education and Research and the Hessian Ministry of Higher Education, Research, Science and the Arts within their joint support of the National Research Center for Applied Cybersecurity ATHENE.

This research was conducted in the context of a fellowship at the Center for Advanced Internet Studies (CAIS).

Funded by the European Union (ERC, InterText, 101054961). Views and opinions expressed are however those of the author(s) only and do not necessarily reflect those of the European Union or the European Research Council. Neither the European Union nor the granting authority can be held responsible for them.

\bibliography{sources}
\bibliographystyle{acl_natbib}

\newpage
\appendix

\section{Application details}
\label{app:appdetails}

\begin{figure}[h]
  \centering
  \includegraphics[width=\linewidth]{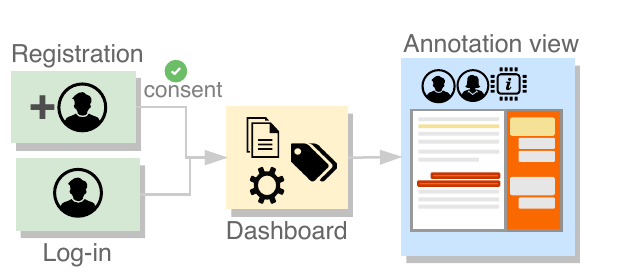}
  
  \caption{User journey in CARE}
  \label{fig:uj}
\end{figure}

\begin{figure*}[h]
  \centering
  \includegraphics[width=\textwidth]{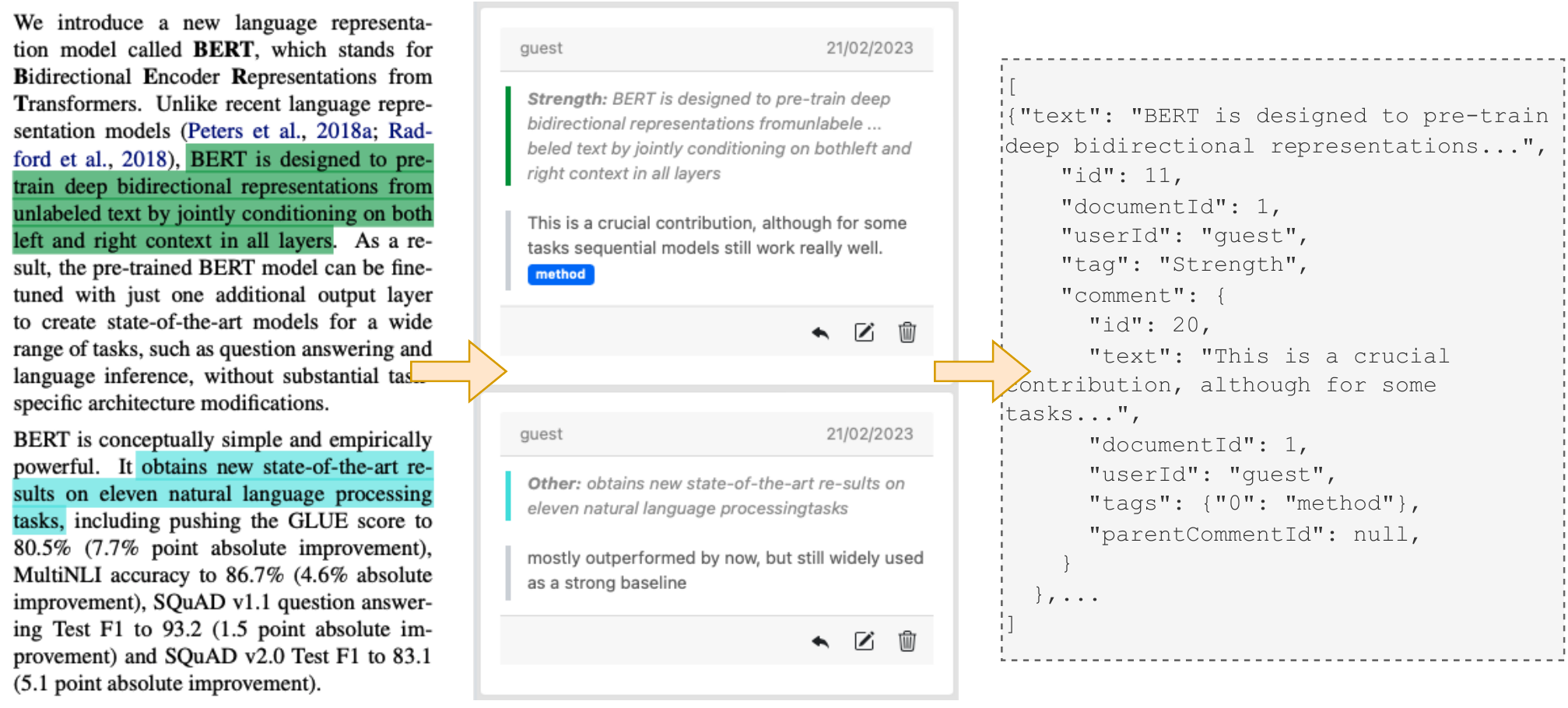}
  
  \caption{Data export example from highlights to annotations in the sidebar, to export JSON.}
  \label{fig:dex}
\end{figure*}

\begin{figure*}[t]
  \centering
  \includegraphics[width=0.8\textwidth]{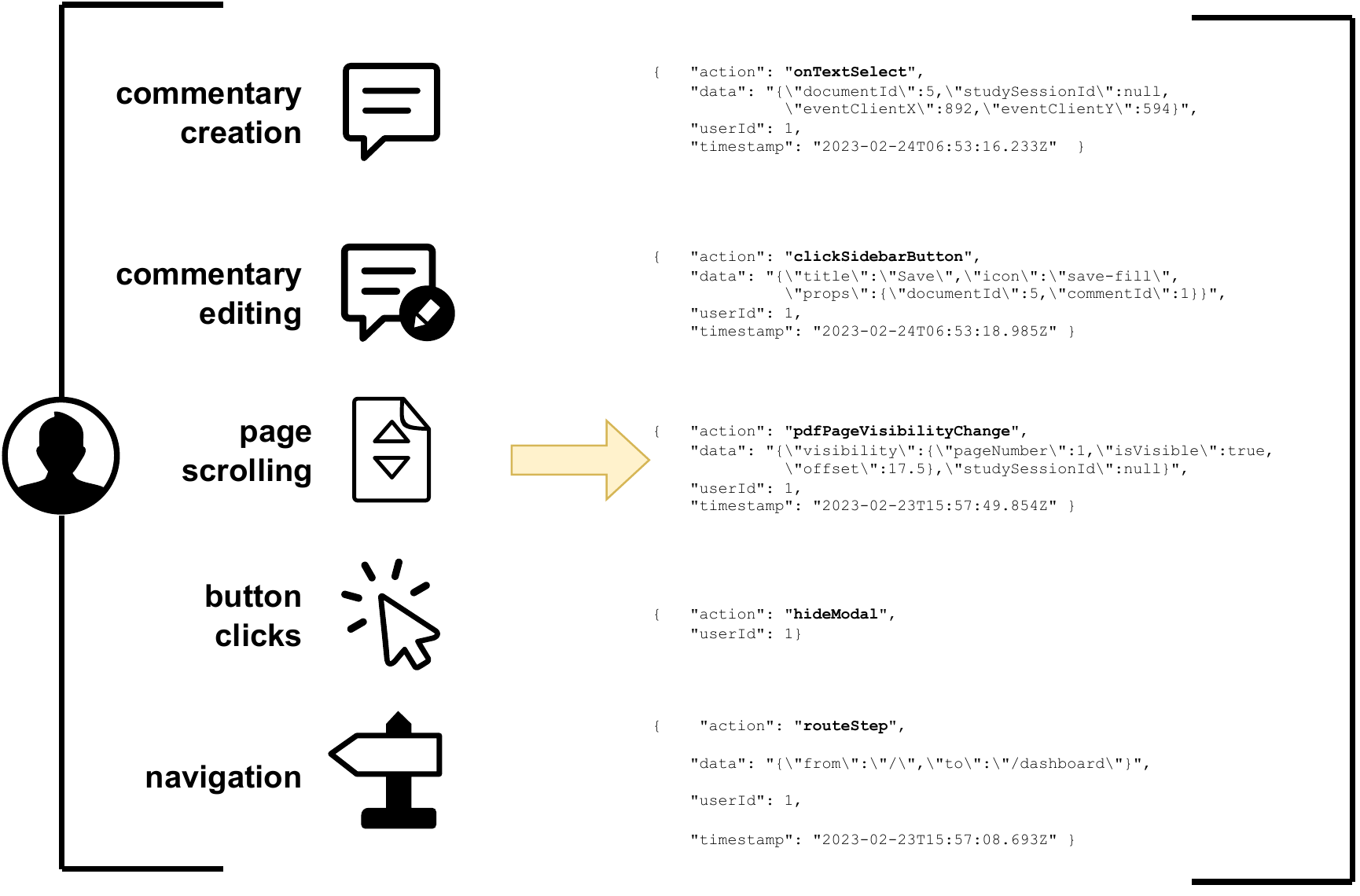}
  
  \caption{Behavioral user data examples captured and exported as JSON objects.}
  \label{fig:statex}
\end{figure*}

\paragraph{User journey} Figure \ref{fig:uj} illustrates a typical user journey for the reader using CARE. It starts with log-in or registration during which consent and licensing forms are submitted. Afterwards, the user is presented with a dashboard where they can manage documents, label sets and additional settings and export inline commentary and behavioral data. Each document can be opened for reading via the annotation component, where multiple users can annotate the document by leaving inline commentaries organized in a sidebar which also serves as an interface for AI assistance.

\paragraph{Export data format} Figure \ref{fig:dex} provides an example of the data export functionality: all annotations, comments and discussion threads created by the readers can be directly exported as an easy-to-use JSON. Note that the information presented in the export is also the information available to NLP assistance models to make their predictions.

\paragraph{Behavioral data}
Figure \ref{fig:statex} provides examples of behavioral data that is captured within CARE and exported as JSON objects. Each action is associated with a unique type, meta-data, user information and a timestamp. The captured user interactions include the creation of inline commentary, editing of the same, page scrolling, clicks on important buttons and navigation within the tool.

\section{User Study Details} \label{as:case-study}
\begin{figure}[t]
  \centering
  \includegraphics[width=\linewidth]{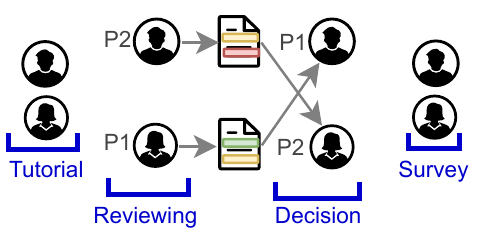}
  \caption{User study setup: Split into two groups after a brief tutorial, the participants review a paper, exchange reviews and make an acceptance decision for the other paper, and participate in the survey.}
  \label{afig:userstudy}
\end{figure}

This section provides extensive details on the user study setup and results. To recap, the participants were instructed to use CARE for leaving inline commentaries on a manuscript with the purpose of assessing the manuscript's quality and scientific merit, similar to the critical reading process that takes place during scholarly peer review. Participants were split into two groups that reviewed one manuscript each. Participants subsequently exchanged the reviewed manuscripts and used the provided inline annotations to decide whether a manuscript should be accepted or rejected, similar to traditional peer review, and surveyed. Figure \ref{afig:userstudy} summarizes the study design. The papers considered were "Academia's Big Five" \cite{paper1_us} (P1) and "The Unhappy Postdoc" \cite{paper2_us} (P2), both in their first version submitted to F1000 Research.

\paragraph{User Study Context}
The user study was implemented as a workshop on 25 August 2022 within the Center for Advanced Internet Studies (CAIS)\footnote{\url{https://www.cais-research.de}}. CAIS is an interdisciplinary research institute in Bochum, Germany, that focuses on the social opportunities and challenges of the digital transformation. Research is conducted in longer-term research programs, as well as by fellows and working groups who are invited to the institute to pursue their own projects. The scientific focus lies on the interface between social sciences, humanities and computer sciences.

\subsection{Participant Pool} \label{ass:pool}
The participant pool for the user study consisted of $11$ CAIS members attending the workshop either virtually ($2$ participants) or in person ($9$ participants). No selection criterion was applied to the voluntary participant pool. %
To ensure the privacy of the participants, we report accumulated frequencies for appropriate value intervals in the following paragraphs.

\paragraph{Demographics} 
Of this participant pool five ($45\%$) identified as women, five ($45\%$) as men and one preferred not to share this information. 
Around $30\%$ of participants report an age below $40$, while the majority of participants lie in the $40-49$ ($45\%$) age range. The rest of the participants ($25\%$) either lie in the age group above $50$ or did not report their age. The majority of participants lived and worked in Germany ($80\%$).
We deem the given sample as sufficiently diverse for the purpose of this study, as it covers various age groups and shows nearly balanced genders. However, the age group below forty is under-represented, which might have an influence on the study results, as this particular group might show higher digital affinity. 
Follow-up studies are required to confirm our findings, where a focus on lower age groups and more diverse nationalities should be considered to account for cultural differences of the partially subjective peer review assessment process.

\paragraph{Academic Background}
At the moment of the study, more than $60\%$ of the participants were at a post-doctoral or professorial level in their careers, ensuring an adequate level of expertise and experience in scholarly text work. The participants came from diverse academic backgrounds including social studies, philosophy, law, natural language processing and literary studies. The vast majority of participants ($90\%$) had no computer science background.

\paragraph{Reviewing Expertise}
In an independent pre-study survey among the CAIS members, we confirmed that English language papers and reviews are the predominant form of scientific communication in their respective fields, suggesting adequate language proficiency of the participants during the study.

Roughly $64\%$ of the participants personally reviewed more than one paper in the past year; only two participants reviewed no papers during their career so-far (zero reviews in the past five years). On average the participants reviewed roughly three papers per year. Apart from the prevalent high academic seniority, these numbers generally suggest deep expertise in the task of peer review, while at the same time the study includes participants with little to no reviewing experience.

\subsection{Post-study Survey} \label{ass:usability}
The participants were asked to fill out the post-study questionnaire directly after the user study. Each participant responded to the web form individually and privately. We ensured the right of erasure under GDPR regulations\footnote{\url{https://gdpr-info.eu/art-17-gdpr}} and hosted the questionnaire and resulting data exclusively on EU servers. The questionnaire contained in total $35$ items structured into the sections demographics and experience and usability.

\paragraph{Quantitative Results}
The usability section consists of five general usability questions answered on a seven-point scale ranging from "Strongly disagree" (1) to "Strongly agree" (7), as well as free form questions about missing features and feedback about specific design choices.
Figure \ref{afig:full_questionaire} shows the answer distribution on the usability questionnaire. We asked participants to rate the overall experience using \toolname{}, the speed of usage, the ease of finding information, the comprehensiveness of features and the utility of the sidebar.

\paragraph{Qualitative Results}
Further on, we asked the participants whether they would prefer different orderings of the comments in the sidebar, where the default during the study was an ordering by text position. While this default is perceived as useful ($36\%$), the option for changing the comment order or other grouping strategies are of interest to the users -- especially in the decision making phase based on the inline comments of a reviewer. Subsequently, we asked users to highlight which features they missed or could think of to streamline their inline peer review. Most requests were directed towards performing a full peer review based off the inline commentary, e.g. providing notes to editors, providing ratings, having the reviewing guidelines integrated in the interface, etc. Further suggested features that were more focused on the actual highlighting and commenting aspects rather than inline peer review, comprised of more extensive PDF viewer features, like zooming, and improved highlighting features, e.g. sentence-boundary aware highlighting, figure selection, and cross-linking of commentary.

\subsection{Behavioral Data} \label{ass:behavior}
In this section we report on the detailed results of the behavioral data tracking during the user study. Besides showcasing the behavioral data tracking capabilities of \toolname{}, we intend to collect insights into usage patterns of the tool, as well as establishing a deeper understanding of the use-case of assisting reviewers during reading.

\paragraph{Task Timing}
We consider several timing metrics to measure the ease of usage, as well as the task difficulty. 

First, we measure the time-to-completion, starting with the users accessing the document and ending with them submitting their inline review. The median time-to-completion amounts to $37.82$min (just below the provided time limit), with a high standard deviation of roughly $13$min. Except for two outliers requiring below $15$min, this suggests most people did use and require the full time interval to perform their inline peer review.

Second, we measure the time passed before the interaction with a feature of \toolname{} was registered. This includes text selections for highlights, scrolling to a new page, or creating a comment in the sidebar. We employ this metric as an indicator for the bandwidth of the perceived user interface complexity. In fact, we see that on median $1.28$min pass before the first interaction, while again showing high standard deviation of $50$s. The high variance and relatively long median time before the first interaction suggest that some participants were still familiarising with the study instructions while already having accessed the document. This shows one limitation of the behavioral tracking implemented in \toolname{} so-far: while the behavioral data logging is non-obstructive to the user experience, unlike e.g. eye-tracking devices in laboratory scenarios, off-screen activities such as breaks cannot be detected reliably. 

\paragraph{Reading and Inline Commentary}
We consider two metrics to analyze the participants' focus of attention during the reading process. As the first metric, we consider the time of inline commentary creation relative to the total task time, to quantify whether participants create annotations throughout the reading process or detached before or after reading.
For an inline commentary $x$ created at $t_c$ we define the \textit{reltime} relative to the user's time of entering the document $t_e$ and the time of leaving the document $t_l$ as:
$$
\textit{reltime}(x) = \frac{t_c(x) - t_e}{t_l - t_e}
$$

\begin{figure}[t]
    \centering
    \includegraphics[width=0.98\linewidth]{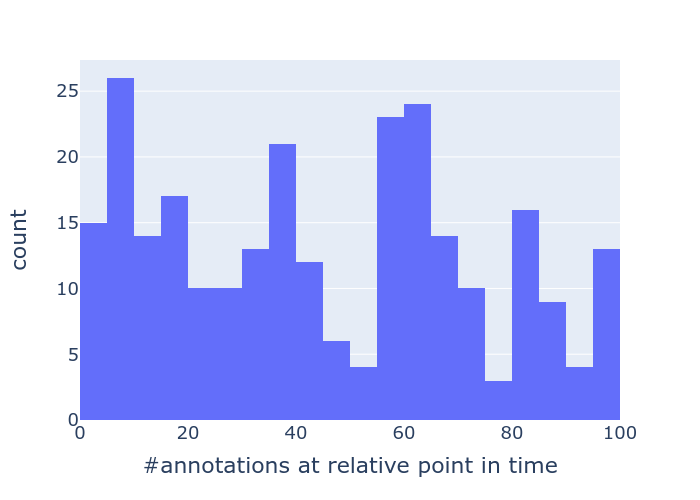}
    \caption{Histogram of the distribution of annotations across time relative to the user's task timing. We accumulate across users.}
    \label{afig:anno_times}
\end{figure}

Figure \ref{afig:anno_times} shows the distribution of relative inline commentary timings across participants. Apparently the participants create annotations throughout the whole annotation process, with a light dip at $50\%$, i.e. after half of the time to completion. These measurements do not suggest that making inline commentary is decoupled from the actual reading process, instead \toolname{} seems to support regular highlighting and note-taking habits while reading.

Turning to the second metric, we compute the time elapsed while viewing a page during the study. We compute the relative reading time per participant and page, considering the two papers in isolation. To estimate the relative time spent per page, we measure the time deltas between two subsequent page view events, indicating that a PDF page has been rendered on the participants screen, and normalize by the total task time. While this metric is a sufficient approximation for the purpose of assessing the overall reading coverage throughout the document, the measurements on page level instead of scrolling positions limit fine-grained claims about the reading position of a user.

\begin{figure}[t]

    \centering
    \includegraphics[width=0.98\linewidth]{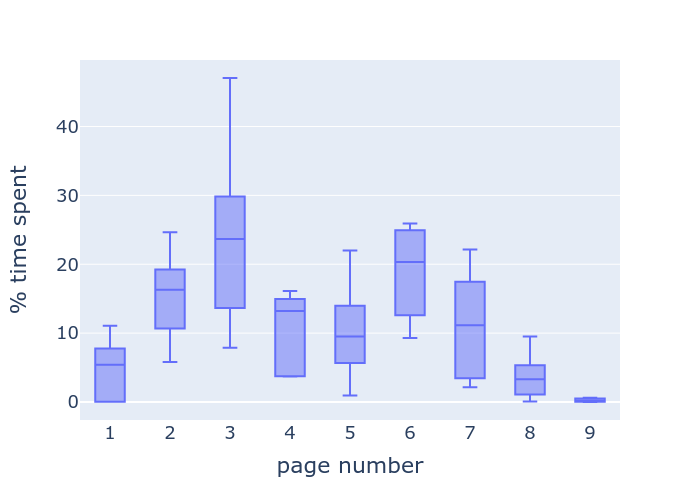}
    \centering
    \includegraphics[width=0.98\linewidth]{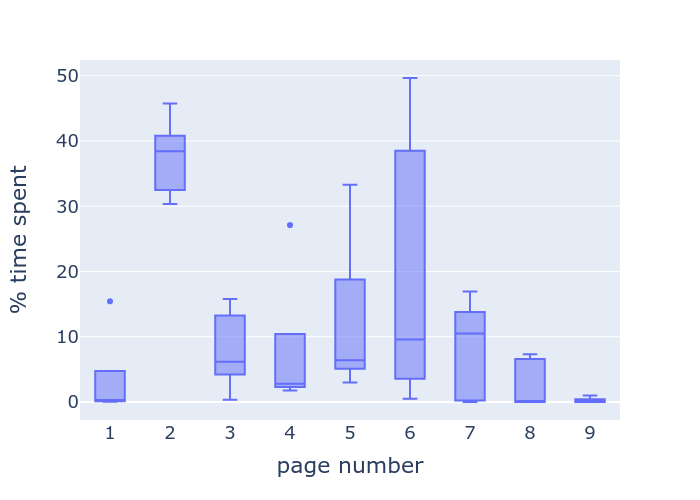}
\caption{Relative reading time per page for papers P1 (top) and P2 (bottom)}
    \label{afig:read_times}
\end{figure}

Figure \ref{afig:read_times} shows the median reading times per page of the users for the two papers in isolation. For both papers, the reading times have a similar "M" shape, where the least amount of time is spent on the very first page, the middle part of the paper and the final pages. For P2 we observe a consistent peak on page two containing the main part of the introduction and, with high variance, page six including the discussion and a central figure of the article. For P1 individual page reading times are less pronounced, but we see peaks on page three (including a large table) and the pages five and six consisting of a long body of text explaining the core contribution (a taxonomy) of the paper.

In the given user study setting, the page viewing times may reveal the parts of the paper that received most scrutiny during reading and commenting, as well as an estimate of the coverage of all paper aspects by the participants. For instance, we see that the bibliography has not been analyzed in detail by any of the participants. In general scenarios, the page viewing times may reveal places of interest in a document or indicate passage that require more effort to process during reading.

\end{document}